\pdfoutput=1
\documentclass[11pt]{article}

\usepackage[]{ACL2023}

\usepackage{times}
\usepackage{latexsym}
\usepackage{amsmath}
\usepackage{booktabs}
\usepackage{enumitem}
\usepackage{tcolorbox}
\setlist{leftmargin=1mm}
\usepackage{pifont}
\usepackage{setspace}

\usepackage{multirow}
\usepackage{longtable}

\usepackage[T1]{fontenc}
\usepackage[utf8]{inputenc}

\usepackage{microtype}

\usepackage{inconsolata}

\title{Beyond Text: Leveraging Multi-Task Learning and Cognitive Appraisal Theory for Post-Purchase Intention Analysis}
\author{Gerard Christopher Yeo$^{1}$, Shaz Furniturewala$^{2}$, {Kokil Jaidka}$^{3}$ \\
$^{1}$Institute of Data Science, National University of Singapore, \\ $^{2}$Birla Institute of Technology and Science, \\ $^{3}$NUS Centre for Trusted Internet and Community, National University of Singapore \\
\texttt{e0545159@u.nus.edu}, \texttt{mohammadshaz529@gmail.com}, \texttt{jaidka@nus.edu.sg}}

\begin{document}
\maketitle
\begin{abstract}
Supervised machine-learning models for predicting user behavior offer a challenging classification problem with lower average prediction performance scores than other text classification tasks. This study evaluates multi-task learning frameworks grounded in Cognitive Appraisal Theory to predict user behavior as a function of users' self-expression and psychological attributes. Our experiments show that users' language and traits improve predictions above and beyond models predicting only from text. Our findings highlight the importance of integrating psychological constructs into NLP to enhance the understanding and prediction of user actions. We close with a discussion of the implications for future applications of large language models for computational psychology.
\end{abstract}

% edit abstract

\section{Introduction}
Natural language processing (NLP) tasks involve predicting outcomes from text, ranging from the implicit attributes of text to the subsequent behavior of the author or the reader. Recent research suggests that user-level features can carry more task-related information than the text itself~\cite{lynn2019tweet}, but these experiments have been conducted in a limited scope. Other studies have explored how the linguistic characteristics of text, such as its politeness or the use of discursive markers, may predict subsequent user behavior~\cite{danescu-niculescu-mizil_computational_2013,niculae2015linguistic}. Yet, these studies offer unimodal perspectives of users through the text they author and lack rich annotations of other user attributes, such as their cognitive and psychological traits. Such data would be especially useful in applied NLP tasks, such as in the context of online reviews, to better contextualize and predict outcomes related to purchase behavior and product recommendations.

In this study, we focus on \textbf{Cognitive Appraisal Theory}, one of the primary theoretical frameworks in Psychology to understand emotional experiences and how they are elicited (antecedents). Central to Cognitive Appraisal Theory is the proposition that emotions are not merely spontaneous reactions but are the result of intricate cognitive evaluations conducted across multiple dimensions of psychological motivation that are of personal significance to one's well-being, as discussed by seminal works in the field \cite{ lazarus1984stress, ortony2022cognitive, scherer2001appraisal, smith1985patterns}. People interpret— or appraise— situations along various dimensions, and the specific manner in which people appraise their situations characterizes the particular emotions they feel. For example, if a consumer evaluates a restaurant experience as slow (goal inconduciveness), the server was specifically being rude to them (unfair), and blames the waiter for such an experience (accountability-other), then the consumer might feel an emotion like \emph{anger}. Our empirical investigation specifically targets the nuances of purchase behavior, guided by a focus on two critical dimensions as illuminated by Cognitive Appraisal Theory:
\vspace{-3mm}
\begin{itemize}[noitemsep]
    \item \textbf{Cognitive appraisals}: The multifaceted evaluative processes through which consumers engage with and interpret their interactions with products, including, but not limited to, the novelty and pleasantness of the consumer-product encounter \cite{yeo2023meta}.
    \item \textbf{Emotions}: The range of emotions consumers may experience during product usage. Emotions such as anger and disappointment are pivotal, as they color the immediate consumer experience and influence subsequent behaviors and attitudes towards the product \cite{ruth2002linking}.
\end{itemize}

\textbf{Setup and Motivation:} This study predicts post-purchase behavior as the outcome of emotions and their antecedents. Prior work has reported that the myriad of emotions experienced by consumers interacting with a product/service \cite{richins1997measuring} can influence post-consumption behaviors (PCB) like future purchases and likelihood to promote the product to others \cite{folkes1987field, lerner2015emotion, nyer1997study, watson2007causes}. Although previous studies have demonstrated that language models capture emotionally relevant features \cite{acheampong2021transformer, deng2021survey}, these studies do not relate such features to other relevant psychological traits such as cognitive appraisals in the understanding of user behavior. Modeling cognitive appraisals and emotions in language models not only aids in predicting behavioral intentions but also explains \emph{why} people have different behavioral intentions after interacting with a product or service. For example, if a person does not want to recommend the product, this could be attributed to appraisals such as low goal-conduciveness or unfairness and emotions such as disappointment and anger. 

% Therefore, for better or worse, emotions drive many meaningful decisions in consumption. 

\begin{figure*}[!ht]
\center
\includegraphics[width=0.6\textwidth, scale=0.5]{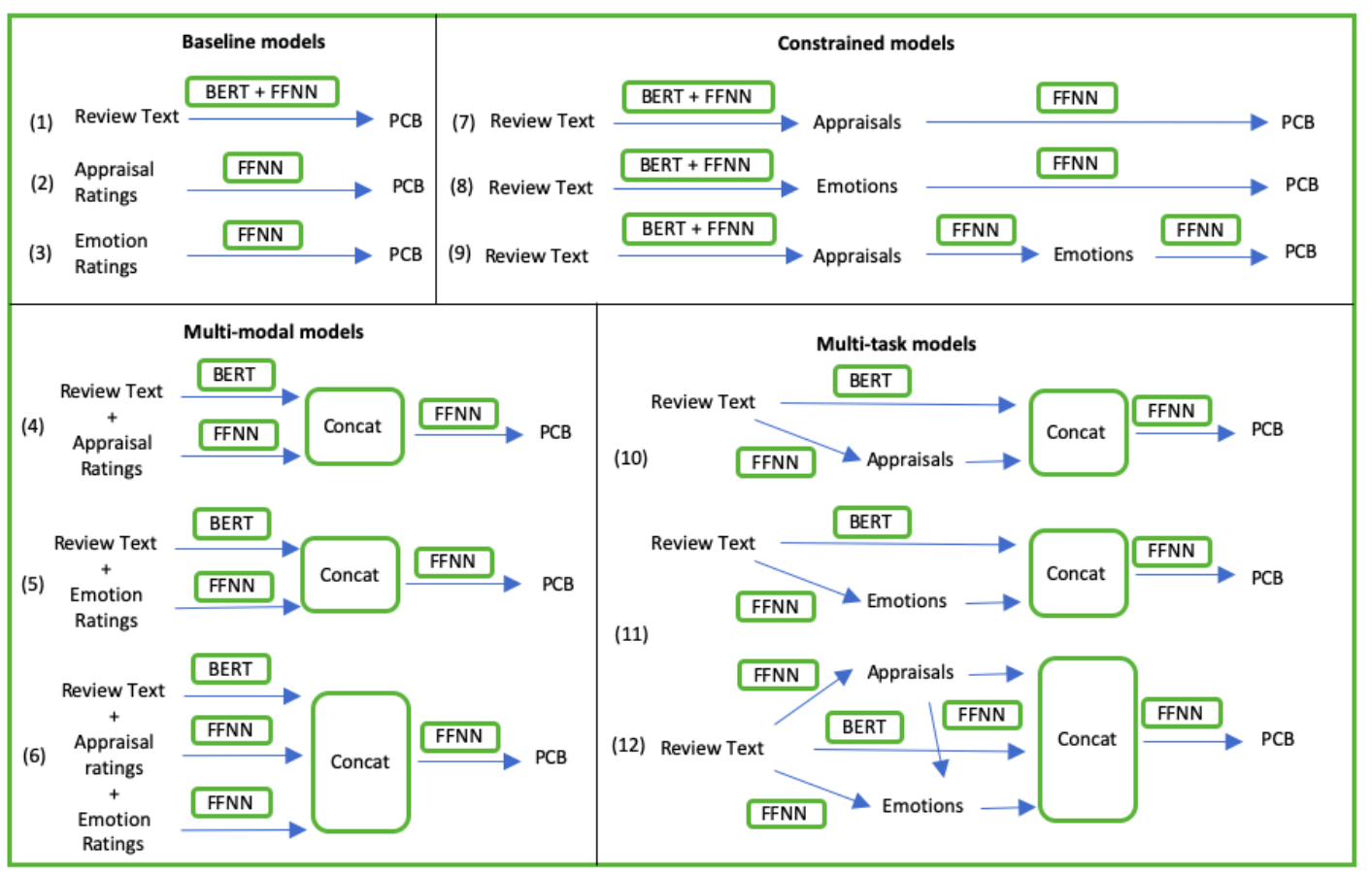} 
\caption{Models implemented in our study. Model (12) is the theoretical model.}
\label{fig:models}
\end{figure*}

We evaluate a series of multi-modal and multi-task learning setups that apply Cognitive Appraisal Theory, as reported in Figure~\ref{fig:models}. The models proposed here are not constructed by merely combining additional information such as emotion and cognitive appraisals but based on theoretical proposals on how such information is related to one another, which is something that previous research has not done in the context of behavioral outcomes \cite{liu2023psyam}. The following are our contributions:
\vspace{-3mm}
\begin{itemize}[noitemsep]
    \item  A multi-task learning framework incorporating emotional and cognitive appraisal variables in a theoretical manner to predict PCB.
\item An exploration of the empirical association of PCB with cognitive appraisals, emotions, and the text authored by the consumer.
       \end{itemize}

%Consumption emotions act as adaptive signals of how the interaction of products/services affects our well-being that subsequently trigger future actions to either promote positive emotions (e.g. repurchasing or promoteing to others) \cite{white2010impact} or reduce negative emotions (e.g. complaint behaviors) \cite{stephens1998don}. For example, if a consumer evaluates a restaurant experience as slow (goal inconduciveness), the server was specifically being rude to them (unfair), and blames the waiter for such an experience (accountability-other), then the consumer might feel an emotion like \emph{anger}. Accordingly, 
%Therefore, to comprehensively model the purchase behavior of consumers, we also need to how it relates to their emotional appraisal; for e.g., \emph{why} does a consumer feel the way they are feeling when using the product?.  %[INSERT the theoretical relationships]

% \section{Method}
% We model PCB intentions in reviews using consumers’ specific emotional experiences and cognitive appraisals. We implemented several computational models, informed by consumer and emotion theory, and tested how cognitive appraisals, and emotional experiences interact with language in reviews to predict PCB intentions. To the best of our knowledge, the current work is the first to study consumer behavior in language from a holistic emotional perspective. Understanding the emotional experiences of consumers and their antecedents (i.e. cognitive appraisals) is important not only for businesses to engineer optimal product/serve usage interactions with customers, but also as an academic endeavour to model consumer behavior in language. Moreover, our study is especially relevant given the rise of social media and Internet platforms providing more opportunities for consumers to share their experiences in the form of reviews.

\section{Dataset and Variables}
We used the PEACE-Reviews Dataset \cite{yeo2023peace}, a dataset of 1,400 author-annotated product reviews describing people’s emotional experiences of using an expensive product/service. To our knowledge, this is the only reviews dataset annotated with a large number of first-person emotional and behavioral intentions variables. Existing emotion text datasets are usually annotated with only a subset of these variables, without the inclusion of any behavioral intentions ratings \cite{scherer1994evidence}. Most importantly, existing emotion text datasets are typically annotated with third-person annotations (i.e., raters rate text written by other people), where such annotations might not correspond to the writers’ first-hand experiences \cite{mohammad2018semeval}. In the PEACE-Reviews Dataset, each review was annotated with first-person emotions, cognitive appraisals, and PCB ratings, which makes the dataset exceptionally relevant in comprehensively modeling consumers’ first-hand emotional experiences and behavior intentions. Our multi-task framework incorporates the following inputs:
\vspace{-3mm}
\begin{itemize}[noitemsep]
    \item \textbf{Review text}. The review text comprises detailed descriptions of consumer-product interactions and specific aspects of the product/service that explain why consumers feel a particular emotion. The mean length of the reviews is 190.2 tokens, which makes them substantively longer than other review datasets \cite{maas2011learning}.
    \item \textbf{Cognitive appraisals}. Each review is annotated with 20 appraisal dimension ratings that measure how consumers evaluate the consumer-product interactions relevant to their emotional experiences \cite{yeo2023peace}. Each dimension is rated on a 7-point Likert scale, assessing the extent to which participants appraised their consumption experience in a particular manner (see Table~\ref{tab:appraisal} in Appendix \ref{apx:appraisalqns}). For example, suppose a participant rated a particular appraisal dimension such as \emph{novelty} as high; it means that they evaluated the product/service usage as a new experience they have never encountered before. 
    \item \textbf{Emotions}. Each review was also annotated on a 7-point Likert scale measuring the intensity for 8 emotions:  \emph{anger}, \emph{disappointment}, \emph{disgust}, \emph{gratitude}, \emph{joy}, \emph{pride}, \emph{regret}, and \emph{surprise}, adapted from the common emotions experienced in a consumption context \cite{richins1997measuring}. Unlike current emotion recognition datasets where each text is labeled with only one emotion \cite{mohammad2018semeval, scherer1994evidence}, the presence of multiple emotion ratings in this dataset is more consistent with real-life situations where consumers typically experience more than one emotion in a consumption context \cite{ruth2002linking}. 
    \item \textbf{Post-consumption behaviors (PCBs)}. These are the primary outcome variables in our study. Two variables in the dataset assessed the likelihood of engaging in different post-consumption behaviors- \emph{intention to repurchase} and \emph{intention to promote}. They are both measured on a 7-point Likert scale. These variables are indicative of whether real actions might be taken in the future \cite{engel1995blackwell}. 
\end{itemize}
\section{Experiments}
See Figure \ref{fig:models} for a visual representation of all models. We fine-tuned the BERT-base model \cite{devlin2018bert} for models requiring input text. We trained feed-forward neural networks (FFNN) for models that require appraisal and emotion ratings as inputs. Since PCBs are rated on a 7-point Likert scale, we segment each rating into low (1-2), moderate (3-5), and high (6-7) and define it as a three-way classification task (see Appendix \ref{apx:pcb_distribution} for the distribution of classes). For multi-task models where appraisals and emotions are outcome variables, we defined a multi-label binary classification task for emotion ratings, where we segment each rating into low (1-4) and high (5-7). This segmentation represents the presence or absence of emotions experienced by the participant in the situation, where only emotions that are felt with high intensity are considered to be present. We define a multi-output classification task for appraisals where we segment each rating into low (1-2), moderate (3-5), and high (6-7). The segmentation of appraisal ratings in this manner is typical in emotion research \cite{smith1987patterns}. We conducted 5 repetitions for each model and obtained the means and standard deviations of the accuracy and F1 scores. Implementation details are in the Appendix \ref{apx:details_model}.

\textbf{Baseline models.} Three models serve as the baselines. We run separate models to predict PCBs for each modality $M_i$, where $M$ = [text, appraisals, emotions]. We would like to observe which modality performs best in predicting PCBs. 

\textbf{Constrained models.} We implemented three models. The first two models use the BERT model fine-tuned on the reviews to predict either the appraisal or emotion ratings, and the resulting embeddings are then used to predict PCBs. The third model uses the BERT model fine-tuned on the reviews to predict appraisals, subsequently uses these appraisal embeddings to predict emotions, and finally uses the resulting emotion embeddings to predict PCBs. According to emotion theory, this follows where appraisals are deemed to be antecedents to emotions, resulting in behaviors \cite{watson2007causes}. They are termed \emph{constrained} because the intermediate variable (appraisals or/and emotions) serves as a bottleneck that reduces the textual dimensions to a much lower dimension in predicting PCBs, compared to directly predicting PCBs from text.

\textbf{Multi-modal models.} We implemented three models. The first two models predicted PCBs using review text + $M_i$, where $M$ = [appraisals, emotions]. The third model predicted PBs from all three modalities. The embeddings of the modalities are concatenated to predict PCBs. This modeling approach is chosen for its capacity to assimilate psychological variables alongside linguistic features. The results allow us to compare whether ratings combined with review text help improve performance predicting PCBs.

\textbf{Multi-task models.} We implemented three models. For the first two models, the review texts are used to predict the PCBs and $R_i$, where $R$ = [appraisals, emotions], simultaneously. Moreover, the embeddings of $R_i$ are used to predict PCBs by concatenating with the text embeddings. The final model, termed 'Theoretical model', uses the review text to predict appraisals, emotions, and PCBs. The resulting embeddings from each modality are then concatenated to predict PCBs. Additionally, we also used the appraisal embeddings to predict emotions. Overall, this model is based on consumer and psychological theories. We would like to validate whether such a computational model consisting of the variables and their theoretical links has predictive utility in the context of language.  

%These models are motivated to provide a multi-task framework that includes necessary variables based on theory to predict PCB end-to-end.

\begin{table}
%\small
\centering
\resizebox{0.5\textwidth}{!}{
\begin{tabular}{p{4cm}cccc}
\hline
& \multicolumn{2}{c}{\textbf{Intent to repurchase}}& \multicolumn{2}{|c}{\textbf{Intent to promote}}\\
\hline
\textbf{Model} & \textbf{Accuracy} & \textbf{F1}& \textbf{Accuracy} & \textbf{F1}\\ \hline
\textbf{Baseline} \\
Text -> PCB & 70.1 (0.29) & 0.61 (0.01) & 72.4 (0.57) & 0.66 (0.01) \\ 
Appraisals -> PCB& 73.4 (0.29) & 0.70 (0.01) & 75.9 (1.05) & 0.74 (0.01)\\ 
Emotions -> PCB  & 67.7 (0.53) & 0.66 (0.01) & 73.3 (0.35) & 0.72 (0.01) \\ \hline
\textbf{Constrained} \\
Text -> Appraisals -> PCB & 69.0 (0.35) & 0.58 (0.01) & 69.6 (0.57) & 0.58 (0.01) \\ 
Text -> Emotions -> PCB & 68.6 (0.45) & 0.58 (0.01) & 69.1 (0.29) & 0.58 (0.01) \\ 
Text -> Appraisals -> Emotions -> PCB  & 67.3 (0.83) & 0.57 (0.01) & 68.3 (0.31) & 0.58 (0.01) \\ \hline
\textbf{Multi-modal} \\
Text + Appraisals -> PCB & 68.0 (0.34) & 0.68 (0.02) & 72.6 (0.97) & 0.70 (0.01) \\ 
Text + Emotions -> PCB & 72.0 (0.21) & 0.66 (0.01) & 70.0 (0.44) & 0.69 (0.02) \\ 
Text + Appraisals + Emotions -> PCB & 72.0 (0.24) & 0.72 (0.01) & 72.0 (0.23) & 0.70 (0.02) \\ \hline
\textbf{Multi-task} \\
Text -> PCB + Appraisals & 69.3 (0.45) & 0.58 (0.01) & 71.7 (0.32) & 0.64 (0.03) \\ 
Text -> PCB + Emotions & 69.1 (0.32) & 0.61 (0.02) & 73.6 (0.58) & 0.67 (0.01\\ \hline
%Text -> PCB + App + Emo & 72.1 & 0.65 \\ \hline
\textbf{Theoretical model} & 69.3 (0.58) & 0.60 (0.02) & 73.4 (0.64) & 0.69 (0.02) \\ \hline

\end{tabular}
}
\caption{Results of three-way (high, medium, low) post-consumption behavior (PCB) classification across models, for intention to promote and intention to repurchase. Values without and within the parentheses represent the means and standard deviations across 5 runs.}
\label{tab:results_main_rec}
\end{table}
\begin{figure*}[!ht]
    \centering
    \includegraphics[width=\textwidth]{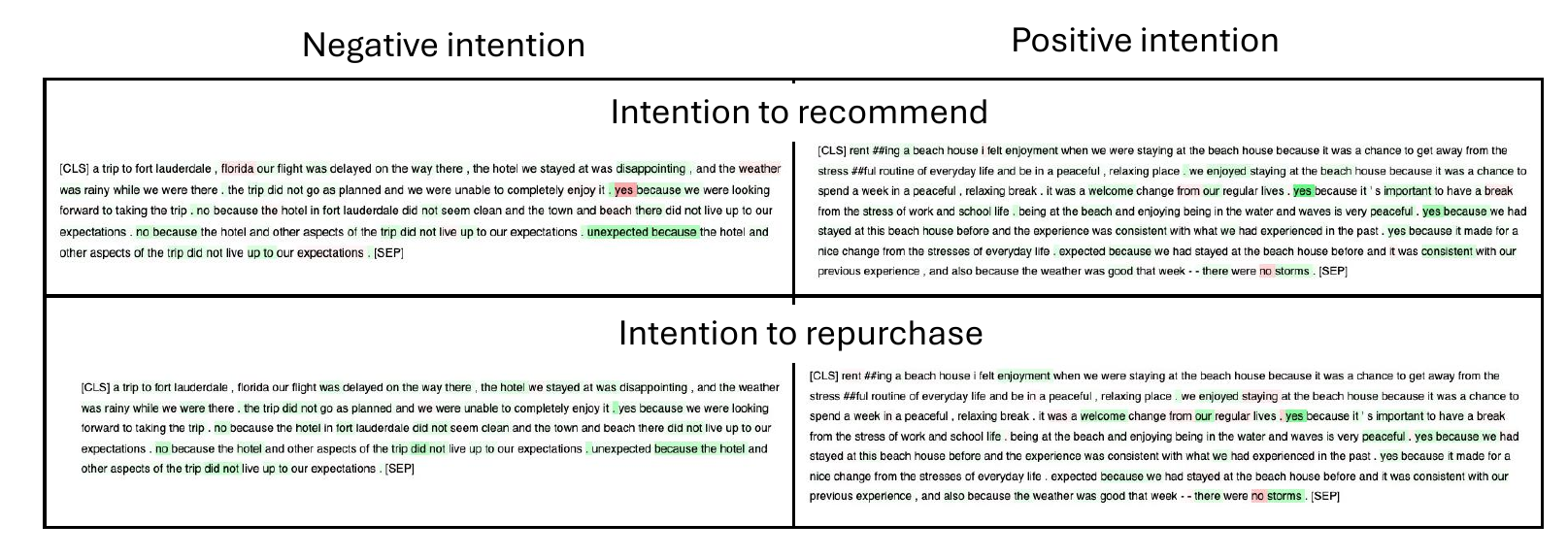} 
    \caption{Word attribution of two samples that scored high and low in PCBs based on the baseline text -> PCB model, respectively.}
   \vspace*{-\baselineskip}
    \label{fig:pos_neg_attri}
\end{figure*}

%Description of the different models and their rationale (why did we choose BERT?)
\section{Results}
%\subsection{Main Findings}
%This section focuses on the results for predicting positive word-of-mouth (Table \ref{tab:results_main_rec}). Please refer to Appendix \ref{apx:intention} for the results on repurchase intentions. 
Table \ref{tab:results_main_rec} presents the results for the different models in predicting the two PCBs. Among the baseline models, models trained directly on appraisals were the most accurate. The Emotions -> PCB model only outperformed the Text -> PCB model in predicting intentions to promote, but not for intentions to repurchase. Despite this, the Text -> PCB model’s performance was still competitive, suggesting that large language models can capture pertinent linguistic features, including those beyond emotional content. 

The poorest results came from constrained models, likely due to the reduction of text embeddings to a lower dimensional appraisal and emotion feature space and resulted in the lost of  information in predicting PCB. 

%The multi-task models achieved comparable results to the baseline, but the theory-based model showed a modest improvement, likely due to its structured integration of appraisal and emotional constructs.

The integration of different modalities (multi-modal models) did not enhance the performance as expected, indicating that unique information from each modality may not be additive for PCB prediction. Nevertheless, some multi-modal models offer a slight edge in accuracy and F1 scores compared to the baseline Text -> PCB model as observed for the results of the Text + Appraisals -> PCB model in predicting intention to promote, and Text + Emotions -> PCB and Text + Appraisals + Emotions models in predicting intent to repurchase.   

For the multi-tasks and theoretical models, for the prediction of intent to promote, the theoretical model and the Text -> PCB + Emotions models outperformed the constrained, multi-modal and baseline Text -> PCB models. In predicting intent to repurchase, the performances of the multi-tasks and theoretical models are similar to the constrained and the baseline Text -> PCB models but did not perform better than the multi-modal models. This suggests that combining appraisals and emotions based on theory might not be optimal in predicting intention to repurchase compared to merely combining the features of the text, appraisals, and emotions.

In general, for intention to promote, the multi-task and theory-informed models performed modestly better than the rest of the models (except for the Appraisals -> PCB model), likely due to their structured integration of appraisal and emotional constructs. However, for intention to repurchase, the two multi-modal models (excluding Text + Appraisals -> PCB) performed the best (except for the Appraisals -> PCB model). Overall, our results affirm that incorporating appraisal and emotional considerations generally enhances PCB prediction and supports the validity of Cognitive Appraisal Theory in informing multi-task learning approaches.

\textbf{Word attributions and explainability.} We implemented the Integrated Gradients method to obtain the word attributions to explain the predictions \cite{sundararajan2017axiomatic}. The visual depictions in Figure~\ref{fig:pos_neg_attri} showcase word attributions corresponding to high and low instances of intentions to promote or repurchase, respectively, predicated upon our baseline Text -> PCB model. The word attributions underscore the integral role of the emotionally-charged lexicon — `enjoyment,' `disappointing' — and cognitive appraisal terms — `unexpected,' `important,' and `consistent' — in influencing the predictive outcomes of our BERT-based model. 

The first two rows indicate that the model's reliance on affective language is pronounced, indicating a robust association between sentiment-laden words and positive intention to promote. In contrast, the word- and phrase- associations with intention to purchase illustrate a less pronounced correlation. We can infer that emotionally resonant words seem more decisive in predicting the intention to promote, while a blend of cognitive appraisal and emotional language informs purchase intentions. This distinction may be crucial for refining the predictive efficacy of sentiment analysis models in consumer behavior contexts.

%Figure \ref{fig:pos_neg_attri_int} shows a sample of high and low intention to recommend and their word attributions using the baseline text-to-PCB model, respectively. Consistent with our prediction results, we observed that emotion words such as `enjoyment' and `disappointing' correlate highly with the predictions. Moreover, words associated with cognitive appraisals, such as `unexpected,' `important,' and `consistent' are also correlated with the predictions. 
%Figure \ref{fig:pos_neg_attri_int} shows the word attribution for purchase intentions. We observed that the model uses words associated with cognitive appraisals (e., `unexpected') and emotions (e.g., `enjoyed') in the prediction. However, the correlation between these appraisals and emotional words is not as strong as the associations for intention to recommend. 
Finally, the figures highlight the errors in how non-cognitive, non-emotional words (e.g., `Florida,' and `hotel') are correlated with PCBs. Overall, our results are consistent with the findings that emotions and appraisals have significant links to PCBs \cite{nyer1998effects}. The analysis of word attributions in our models sheds light on the cognitive processes underpinning specific emotional reactions and behavioral tendencies. Therefore, fine-tuning transformer models with appraisal and emotional variables and identifying linguistic features of such variables can potentially improve the prediction of PCBs. Future studies could implement models that learn these variables simultaneously in a multi-task framework, thereby predicting PCBs. 

\section{Conclusion}
Many NLP tasks focus on predicting user behavior, and enriching text-based models with user and social contexts is increasingly necessary. This work emphasizes the increasingly prominent role of cognitive and emotional signals in behavioral prediction. Consumption emotions act as adaptive signals of how we evaluate how the use of products/services affects our well-being, which subsequently triggers future actions to either promote positive emotions (e.g., repurchasing or promoting to others) \cite{white2010impact} or reduce negative emotions (e.g., complaint behaviors) \cite{stephens1998don}. To our knowledge, the current work is the first to construct models grounded on psychological theory to model real post-consumption decision-making processes, and we find empirical support for these associations. More broadly, our study offers a novel methodological approach to study psychological variables in the context of empirically validating theoretical relationships within review texts—a domain previously unexplored beyond the confines of traditional survey methods. Our work finds variance in the importance of these appraisals across tasks, raising important practical considerations for designing future approaches to behavioral prediction.

\section*{Limitations}
This study used a dataset primarily curated to study emotional responses in review text in the context of using expensive products/services. Although we have established that emotional constructs are important in modeling PCB intentions, one limitation is that the current results might not generalize to other review datasets and contexts. One research direction we would like to pursue is to analyze whether the results from fine-tuning models on the PEACE-Reviews dataset can generalize to other public review datasets with different emotional content, length, contexts, and product/service types. Moreover, since typical review datasets only contain ratings of sentiments and helpfulness, to establish the criterion validity of our models in measuring PCBs, we can estimate the correspondence between predicted PCB scores of our models with other ratings like sentiment and helpfulness. This can further solidify the case that emotion and appraisals are important variables in modeling consumer experiences and behaviors. 

Another limitation is that the dataset only provides ratings for 8 emotional experiences. Although we mentioned that these emotions are typically experienced during consumption, they might not comprehensively capture all emotional experiences \cite{ richins1997measuring}. Despite that, we accounted for the observation that consumers might experience multiple emotions in a situation and also used appraisal dimension ratings to model emotional experiences. Since cognitive appraisal theory posits a one-to-one mapping between appraisal profiles and emotional experiences \cite{ ellsworth2003appraisal}, modeling the 20 appraisal dimensions could mitigate the issue of not comprehensively capturing a wide range of emotional experiences.

\section*{Ethics Statement}
Since we did not collect any data from human subjects but instead used an existing dataset that a review board has reviewed, we do not foresee any potential harm in the methodology of the current study. Moreover, no personal information that could identify individual human participants was in the dataset which can cause privacy issues. 

Extensive literature corroborates the significant impact of cognitive appraisals and emotions on consumer behavior. Our study's objective, to model consumer behavior through emotional variables in review texts, is anchored in a vision of advancing product design and business strategies. Note that the intention of this study is not to manipulate emotional and psychological traits to influence consumer behaviors, but rather to understand and predict consumer behaviors more accurately, thereby contributing valuable insights for an informed decision-making process in business practices. The empirical results and models offered in this study can have potential positive managerial implications such as informing marketing strategies, business decisions, and product engineering. Therefore, users of our models should tailor them to their use cases to aid in understanding consumer behaviors in their specific domain. Furthermore, the current work also adopted the Integrated Gradients method to explain the models' predictions to improve the transparency and interpretability of models to better shape users’ decisions. This ensures that decisions are supported by linguistic features in reviews that have theoretical links with PCBs. 

\textbf{Acknowledgments.} This work is supported by the Ministry of Education, Singapore, under its MOE AcRF TIER 3 Grant (MOE-MOET32022-0001).

\bibliography{anthology,custom}

\begin{thebibliography}{29}
\expandafter\ifx\csname natexlab\endcsname\relax\def\natexlab#1{#1}\fi

\bibitem[{Acheampong et~al.(2021)Acheampong, Nunoo-Mensah, and Chen}]{acheampong2021transformer}
Francisca~Adoma Acheampong, Henry Nunoo-Mensah, and Wenyu Chen. 2021.
\newblock Transformer models for text-based emotion detection: a review of bert-based approaches.
\newblock \emph{Artificial Intelligence Review}, 54(8):5789--5829.

\bibitem[{Danescu-Niculescu-Mizil et~al.(2013)Danescu-Niculescu-Mizil, Sudhof, Jurafsky, Leskovec, and Potts}]{danescu-niculescu-mizil_computational_2013}
Cristian Danescu-Niculescu-Mizil, Moritz Sudhof, Dan Jurafsky, Jure Leskovec, and Christopher Potts. 2013.
\newblock A computational approach to politeness with application to social factors.
\newblock In \emph{Proceedings of the 51st {Annual} {Meeting} of the {Association} for {Computational} {Linguistics} ({Volume} 1: {Long} {Papers})}, pages 250--259, Sofia, Bulgaria. Association for Computational Linguistics.

\bibitem[{Deng and Ren(2021)}]{deng2021survey}
Jiawen Deng and Fuji Ren. 2021.
\newblock A survey of textual emotion recognition and its challenges.
\newblock \emph{IEEE Transactions on Affective Computing}, 14(1):49--67.

\bibitem[{Devlin et~al.(2018)Devlin, Chang, Lee, and Toutanova}]{devlin2018bert}
Jacob Devlin, Ming-Wei Chang, Kenton Lee, and Kristina Toutanova. 2018.
\newblock Bert: Pre-training of deep bidirectional transformers for language understanding.
\newblock \emph{arXiv preprint arXiv:1810.04805}.

\bibitem[{Ellsworth and Scherer(2003)}]{ellsworth2003appraisal}
Phoebe~C Ellsworth and Klaus~R Scherer. 2003.
\newblock Appraisal processes in emotion.

\bibitem[{Engel and Roger(1995)}]{engel1995blackwell}
James~F Engel and D~Roger. 1995.
\newblock Blackwell (1982), consumer behavior.
\newblock \emph{New York: Holt, Renehard, and Winston}.

\bibitem[{Folkes et~al.(1987)Folkes, Koletsky, and Graham}]{folkes1987field}
Valerie~S Folkes, Susan Koletsky, and John~L Graham. 1987.
\newblock A field study of causal inferences and consumer reaction: the view from the airport.
\newblock \emph{Journal of consumer research}, 13(4):534--539.

\bibitem[{Lazarus and Folkman(1984)}]{lazarus1984stress}
Richard~S Lazarus and Susan Folkman. 1984.
\newblock \emph{Stress, appraisal, and coping}.
\newblock Springer publishing company.

\bibitem[{Lerner et~al.(2015)Lerner, Li, Valdesolo, and Kassam}]{lerner2015emotion}
Jennifer~S Lerner, Ye~Li, Piercarlo Valdesolo, and Karim~S Kassam. 2015.
\newblock Emotion and decision making.
\newblock \emph{Annual review of psychology}, 66:799--823.

\bibitem[{Liu and Jaidka(2023)}]{liu2023psyam}
Xuan Liu and Kokil Jaidka. 2023.
\newblock I am psyam: Modeling happiness with cognitive appraisal dimensions.
\newblock In \emph{Findings of the Association for Computational Linguistics: ACL 2023}, pages 1192--1210.

\bibitem[{Lynn et~al.(2019)Lynn, Giorgi, Balasubramanian, and Schwartz}]{lynn2019tweet}
Veronica Lynn, Salvatore Giorgi, Niranjan Balasubramanian, and H~Andrew Schwartz. 2019.
\newblock Tweet classification without the tweet: An empirical examination of user versus document attributes.
\newblock In \emph{Proceedings of the third workshop on natural language processing and computational social science}, pages 18--28.

\bibitem[{Maas et~al.(2011)Maas, Daly, Pham, Huang, Ng, and Potts}]{maas2011learning}
Andrew Maas, Raymond~E Daly, Peter~T Pham, Dan Huang, Andrew~Y Ng, and Christopher Potts. 2011.
\newblock Learning word vectors for sentiment analysis.
\newblock In \emph{Proceedings of the 49th annual meeting of the association for computational linguistics: Human language technologies}, pages 142--150.

\bibitem[{Mohammad et~al.(2018)Mohammad, Bravo-Marquez, Salameh, and Kiritchenko}]{mohammad2018semeval}
Saif Mohammad, Felipe Bravo-Marquez, Mohammad Salameh, and Svetlana Kiritchenko. 2018.
\newblock Semeval-2018 task 1: Affect in tweets.
\newblock In \emph{Proceedings of the 12th international workshop on semantic evaluation}, pages 1--17.

\bibitem[{Niculae et~al.(2015)Niculae, Kumar, Boyd-Graber, and Danescu-Niculescu-Mizil}]{niculae2015linguistic}
Vlad Niculae, Srijan Kumar, Jordan Boyd-Graber, and Cristian Danescu-Niculescu-Mizil. 2015.
\newblock Linguistic harbingers of betrayal: A case study on an online strategy game.
\newblock In \emph{Proceedings of the 53rd Annual Meeting of the Association for Computational Linguistics and the 7th International Joint Conference on Natural Language Processing (Volume 1: Long Papers)}, pages 1650--1659.

\bibitem[{Nyer(1997)}]{nyer1997study}
Prashanth~U Nyer. 1997.
\newblock A study of the relationships between cognitive appraisals and consumption emotions.
\newblock \emph{Journal of the Academy of Marketing Science}, 25(4):296--304.

\bibitem[{Nyer(1998)}]{nyer1998effects}
Prashanth~U Nyer. 1998.
\newblock The effects of satisfaction and consumption emotion on actual purchasing behavior: An exploratory study.
\newblock \emph{The Journal of Consumer Satisfaction, Dissatisfaction and Complaining Behavior}, 11:62--68.

\bibitem[{Ortony et~al.(2022)Ortony, Clore, and Collins}]{ortony2022cognitive}
Andrew Ortony, Gerald~L Clore, and Allan Collins. 2022.
\newblock \emph{The cognitive structure of emotions}.
\newblock Cambridge university press.

\bibitem[{Richins(1997)}]{richins1997measuring}
Marsha~L Richins. 1997.
\newblock Measuring emotions in the consumption experience.
\newblock \emph{Journal of consumer research}, 24(2):127--146.

\bibitem[{Ruth et~al.(2002)Ruth, Brunel, and Otnes}]{ruth2002linking}
Julie~A Ruth, Frederic~F Brunel, and Cele~C Otnes. 2002.
\newblock Linking thoughts to feelings: Investigating cognitive appraisals and consumption emotions in a mixed-emotions context.
\newblock \emph{Journal of the Academy of Marketing Science}, 30:44--58.

\bibitem[{Scherer et~al.(2001)Scherer, Schorr, and Johnstone}]{scherer2001appraisal}
Klaus~R Scherer, Angela Schorr, and Tom Johnstone. 2001.
\newblock \emph{Appraisal processes in emotion: Theory, methods, research}.
\newblock Oxford University Press.

\bibitem[{Scherer and Wallbott(1994)}]{scherer1994evidence}
Klaus~R Scherer and Harald~G Wallbott. 1994.
\newblock Evidence for universality and cultural variation of differential emotion response patterning.
\newblock \emph{Journal of personality and social psychology}, 66(2):310.

\bibitem[{Smith and Ellsworth(1985)}]{smith1985patterns}
Craig~A Smith and Phoebe~C Ellsworth. 1985.
\newblock Patterns of cognitive appraisal in emotion.
\newblock \emph{Journal of personality and social psychology}, 48(4):813.

\bibitem[{Smith and Ellsworth(1987)}]{smith1987patterns}
Craig~A Smith and Phoebe~C Ellsworth. 1987.
\newblock Patterns of appraisal and emotion related to taking an exam.
\newblock \emph{Journal of personality and social psychology}, 52(3):475.

\bibitem[{Stephens and Gwinner(1998)}]{stephens1998don}
Nancy Stephens and Kevin~P Gwinner. 1998.
\newblock Why don’t some people complain? a cognitive-emotive process model of consumer complaint behavior.
\newblock \emph{Journal of the Academy of Marketing science}, 26(3):172--189.

\bibitem[{Sundararajan et~al.(2017)Sundararajan, Taly, and Yan}]{sundararajan2017axiomatic}
Mukund Sundararajan, Ankur Taly, and Qiqi Yan. 2017.
\newblock Axiomatic attribution for deep networks.
\newblock In \emph{International conference on machine learning}, pages 3319--3328. PMLR.

\bibitem[{Watson and Spence(2007)}]{watson2007causes}
Lisa Watson and Mark~T Spence. 2007.
\newblock Causes and consequences of emotions on consumer behaviour: A review and integrative cognitive appraisal theory.
\newblock \emph{European Journal of Marketing}, 41(5/6):487--511.

\bibitem[{White(2010)}]{white2010impact}
Christopher~J White. 2010.
\newblock The impact of emotions on service quality, satisfaction, and positive word-of-mouth intentions over time.
\newblock \emph{Journal of marketing management}, 26(5-6):381--394.

\bibitem[{Yeo and Jaidka(2023)}]{yeo2023peace}
Gerard Yeo and Kokil Jaidka. 2023.
\newblock The peace-reviews dataset: Modeling cognitive appraisals in emotion text analysis.
\newblock In \emph{Findings of the Association for Computational Linguistics: EMNLP 2023}, pages 2822--2840.

\bibitem[{Yeo and Ong(2023)}]{yeo2023meta}
Gerard Christopher Zheng~Jie Yeo and Desmond~C Ong. 2023.
\newblock A meta-analytic review of the associations between cognitive appraisals and emotions in cognitive appraisal theory.

\end{thebibliography}
\bibliographystyle{acl_natbib}

\appendix

\section{Measures of Appraisal Dimensions}
\label{apx:appraisalqns}
Table~\ref{tab:appraisal} provides the items measuring the 20 appraisal dimensions in the PEACE-Reviews Dataset. Each item is measured on a 7-point Likert scale, indicating the endorsement of the particular appraisal dimension used in the evaluation of the person-product interaction.

\begin{table}[!h]
 \small
\resizebox{0.5\textwidth}{!}{%
\begin{tabular}{p{3.5cm}p{4.5cm}}
\hline
\textbf{Appraisal} & \textbf{Measure} \\ \hline
Accountability-circumstances & To what extent did you think that circumstances beyond anyone's control were responsible for what was happening in the situation? \\ 

Accountability-other & To what extent did you think that someone else other than you was responsible for what was happening in the situation? \\

Accountability-self & To what extent did you think that you were responsible for what was happening in the situation? \\

Attentional activity & To what extent did you think that you needed to attend to the situation further? \\

Certainty & To what extent did you understand what was happening in the situation? \\

Control-circumstances & To what extent did you think that circumstances beyond anyone's control were controlling what was happening in the situation? \\

Control-other & To what extent did you think that other people were controlling what was happening in the situation? \\

Control-self & To what extent did you think you had control over the situation? \\

Coping potential & To what extent were you able to cope with any negative consequences of the situation? \\

Difficulty & To what extent did you think that the situation was difficult? \\

Effort & To what extent did you think that you needed to exert effort to deal with the situation? \\

Expectedness & To what extent did you expect the situation to occur? \\

External normative significance & To what extent did you think that the situation was consistent with external and social norms? \\

Fairness & To what extent did you think the situation was fair? \\ 

Future expectancy & To what extent did you think that the situation would get worse/better? \\

Goal conduciveness & To what extent was the situation consistent with what you wanted? \\

Goal relevance & To what extent did you think that the situation was relevant to what you wanted? \\

Novelty & To what extent did you think that the situation was familiar? \\ 

Perceived obstacle & To what extent did you think that there were problems that had to be solved before you could get what you wanted? \\

Pleasantness & To what extent did you think that the situation was pleasant? \\
\hline

\end{tabular}
}
\caption{The cognitive appraisal dimensions measured in the PEACE-Review Dataset.}
\label{tab:appraisal}
\end{table}

\section{Distribution of PCB classes}
\label{apx:pcb_distribution}
Table~\ref{tab:pcb_distribution} provides the distribution of the three PCB classes for the whole dataset. 

\begin{table}[!h]

  \small
\resizebox{0.5\textwidth}{!}{%
\begin{tabular}{lccc}
\hline
\textbf{PCB} & \textbf{Low (\%)} & \textbf{Medium (\%)} & \textbf{High (\%)}\\ \hline
Intent to repurchase & 36.6 & 21.3 & 42.1\\

Intent to promote & 32.3 & 24.1 & 43.6\\ \hline

\end{tabular}
}
\caption{Distribution of post-consumption behavioral (PCB) intentions in terms of the low (1-2), moderate (3-5), and high (6-7) classes for the whole dataset used in the 3-way PCB classification task.}
\label{tab:pcb_distribution}
\end{table}

\section{Model Details and Implementation}
\label{apx:details_model}
We split the dataset up into training, validation, and test sets using 80:10:10 configuration. Since the primary task of predicting PCB is a three-way classification task, we implemented cross-entropy loss for all models to predict PCBs. We used binary cross-entropy loss for appraisal and emotion prediction in multi-task models. Adam optimizer was used with a learning rate of 0.00001. A linear scheduler was also implemented during training. This setting was applied in all models. All models consisting of text inputs are trained for 10 epochs. We found that the performance is stagnant and the fine-tuned BERT models overfit after 10 epochs. On the other hand, models that only use appraisal/emotion ratings are trained from scratch for 2000 epochs, where overfitting occurs after. We implemented separate models for the two PCB variables- a) intention to promote, and b) intention to repurchase. For evaluation, we used the accuracy and the weighted F1 scores. 

\textbf{Baseline models.} For the text -> PCB model, we fine-tuned BERT on the dataset and added a FFNN at the last layer to predict PCB. For the appraisal/emotion -> PCB models, we trained a neural network that has 3 layers of 1024, 512, and 3 nodes, respectively. 

\textbf{Constrained models.} For the Text -> Appraisals/Emotions -> PCB models, the embeddings are obtained after passing to the BERT model. These embeddings are then fed to a FFNN to predict the appraisals/emotions. After which it goes through 3 layers of FFNN of 1024, 512, and 3 neurons, respectively. For the Text -> Appraisals -> Emotions -> PCB model, the appraisal dimensions obtained after passing through the BERT model are fed into a FFNN of 2 layers of 512, and 8, respectively. This 8-dimensional emotion vector is then fed into another FFNN which has 3 layers of 1024, 512, and 3, respectively. 

\textbf{Multi-modal models.} The model of each modality was trained separately to predict PCB. After which, the second-to-last layers (excluding the final FFNN layer) of the models are concatenated and passed through a FFNN of 3 layers of 1024, 512, and 3 nodes, respectively. 

\textbf{Multi-task and theoretical models.} For the two multi-task models that predict appraisals/emotions and PCB, the embeddings of the text reviews after passing through the BERT model are used to predict either the appraisal or emotions through a 1-layer FFNN. After which the result is concatenated with the BERT embeddings and feed through 2 FFNN of 512 and 3 neurons to predict PCB. For the theoretical model, the BERT embeddings are used to first predict the appraisals through a 1-layer FFNN. After which, the resulting embeddings go through 2 FFNN of 512 and 8 neurons to predict the emotions. The BERT, appraisal, and emotion embeddings are then concatenated and feed through 2 FFNN of 512 and 3 neurons to predict PCB.

\end{document}